\documentclass[11pt]{article}

\usepackage[utf8]{inputenc}
\usepackage[T1]{fontenc}
\usepackage{geometry}
\usepackage{amsmath, amssymb}
\usepackage{booktabs}
\usepackage{array}
\usepackage{enumitem}
\usepackage{hyperref}
\usepackage[numbers,sort&compress]{natbib}
\usepackage{lineno}
\usepackage{xcolor}
\usepackage{graphicx}
\usepackage{authblk}

\hypersetup{
  colorlinks=true,
  linkcolor=blue!70!black,
  citecolor=blue!70!black,
  urlcolor=blue!70!black
}

\title{\textbf{Infrastructure-Centric World Models: Bridging Temporal Depth and Spatial Breadth for Roadside Perception}}

\author[1]{Siyuan Meng \thanks{Corresponding author. Email: \texttt{siyuanmeng@umass.edu}}}
\author[1]{Chengbo Ai}
\affil[1]{Department of Civil and Environmental Engineering, University of Massachusetts Amherst}

\date{}

\begin{document}

\maketitle

% ============================================================
\begin{abstract}
% ============================================================
World models,generative AI systems that learn to simulate how environments evolve,have emerged as a cornerstone of next-generation autonomous driving, exemplified by the Waymo World Model, NVIDIA Cosmos, and a rapidly growing body of academic work (HERMES, ResWorld, DriveVLA-W0). However, all existing driving world models adopt an \textit{ego-vehicle} perspective, leaving the infrastructure viewpoint entirely unexplored. We argue that infrastructure-centric world models offer a fundamentally different and complementary capability: the bird's-eye, multi-sensor, persistent viewpoint that roadside systems uniquely possess. Central to our thesis is a \textit{fundamental spatio-temporal complementarity}: fixed roadside sensors excel at \textbf{temporal depth},accumulating long-term behavioral distributions including rare safety-critical events at specific locations,while vehicle-borne sensors excel at \textbf{spatial breadth},sampling diverse scenes across large road networks. Neither platform alone can learn a complete model of traffic dynamics. This paper presents a research vision for developing Infrastructure-centric World Models (I-WM) organized in three phases: (I)~generative scene understanding with quality-aware uncertainty propagation, (II)~physics-informed predictive dynamics with multi-agent counterfactual reasoning, and (III)~collaborative world models for vehicle-to-everything (V2X) communication via latent space alignment. We propose a dual-layer architecture,modular, annotation-free perception as a multi-modal data engine feeding end-to-end generative world models,accompanied by a phased multi-modal sensor strategy that progresses from LiDAR-centric core sensing through 4D radar and signal phase data to forward-looking modalities including event cameras and environmental sensors. We establish a taxonomy of driving world model paradigms, position I-WM relative to LeCun's JEPA framework, Li Fei-Fei's spatial intelligence vision, and Vision-Language-Action architectures, and introduce the concept of \textbf{Infrastructure VLA (I-VLA)} as a novel unification of roadside perception, natural language commands, and traffic control actions. Our vision builds upon existing multi-LiDAR perception pipelines (FRGB3D, MulDet3D, MulTrack3D) and identifies concrete open-source foundations (DynamicCity, OccWorld, CarDreamer) for each phase, providing an actionable path toward intelligent infrastructure that does not merely observe traffic but understands and anticipates it.
\end{abstract}


% ============================================================
\section{Introduction}
\label{sec:intro}
% ============================================================

Current roadside perception systems primarily focus on instantaneous object detection and tracking. However, they lack a deep understanding of the ``rules of the world'',such as physical constraints, causal relationships, and intent prediction. Meanwhile, industry leaders such as Waymo have recently introduced vehicle-centric world models built upon foundation models like Google DeepMind's Genie~3, demonstrating the power of generative simulation for autonomous driving safety validation~\cite{waymo2026worldmodel}. Concurrently, NVIDIA's Cosmos platform~\cite{nvidia2025cosmos} has established open-weight world foundation models for physical AI, while a rapidly growing body of work,including HERMES~\cite{zhou2025hermes}, ResWorld~\cite{zhang2026resworld}, and DriveVLA-W0~\cite{li2025drivevlaw0},continues to push the frontier of vehicle-centric driving world models.

This paper proposes a 5-year research trajectory to develop \textbf{Infrastructure-centric World Models (I-WM)} that offer a fundamentally different and complementary perspective: the \textit{bird's-eye, multi-sensor, persistent} viewpoint that roadside systems uniquely possess. Unlike ego-vehicle world models that reason about the immediate surroundings of a single agent, I-WM reasons about the \textit{entire traffic ecosystem},enabling predictive analytics, proactive safety interventions, and cognitive V2X collaboration. Central to our thesis is a \textbf{fundamental spatio-temporal complementarity}: roadside systems excel at \textit{temporal depth},accumulating long-term behavioral distributions at fixed locations,while vehicle-based systems excel at \textit{spatial breadth},sampling diverse scenes across large road networks. The I-WM framework is designed to harness this complementarity as a first-class design principle.

Architecturally, we advocate a \textbf{dual-layer design} that separates concerns between perception and world modeling. The \textit{perception layer} employs modular, unsupervised methods,background modeling (FRGB3D), multi-objective detection (MulDet3D), and consensus-based tracking (MulTrack3D),that operate without manual annotation and can be rapidly deployed at new sites. These serve as a structured \textit{data engine}, providing the world model with high-quality, uncertainty-annotated observations (3D detections, trajectories, static scene priors). The \textit{world model layer} then builds upon these structured inputs using end-to-end generative architectures,diffusion-based 4D scene generation, autoregressive occupancy forecasting, and self-supervised latent dynamics,to learn scene evolution, counterfactual reasoning, and future prediction. This separation is deliberate: roadside deployment demands annotation-free perception (since each new intersection lacks labeled data), while world models benefit from end-to-end learning over the structured representations that the perception layer produces. In the longer term, we envision a fully end-to-end roadside foundation model that unifies both layers, but the dual-layer approach provides an actionable near-term path that leverages existing validated components.

% ============================================================
\section{Industry Context: Why Infrastructure-Centric?}
\label{sec:context}
% ============================================================

The recent release of the Waymo World Model (February 2026) marks a milestone for vehicle-centric generative simulation. Built on Genie~3, it generates photorealistic camera and LiDAR data for rare edge-case scenarios, supports counterfactual reasoning via driving action control, and enables language-conditioned scene editing. However, Waymo's approach is fundamentally \textit{ego-centric},it models the world from a single vehicle's perspective.

\textbf{Infrastructure-centric and vehicle-centric world models are complementary, not competing.} The roadside ``God's-eye view'' resolves occlusions invisible to any single vehicle, provides persistent monitoring across all traffic participants, and enables system-level optimization (e.g., signal control) that ego-centric models cannot achieve. Table~\ref{tab:comparison} summarizes these differences.

\renewcommand{\arraystretch}{1.3}
\begin{table}[t]
\centering
\caption{Positioning: Vehicle-Centric vs.\ Infrastructure-Centric World Models.}
\label{tab:comparison}
\small
\begin{tabular}{>{\raggedright\arraybackslash}p{3cm} >{\raggedright\arraybackslash}p{5.5cm} >{\raggedright\arraybackslash}p{5.5cm}}
\toprule
\textbf{Dimension} & \textbf{Waymo World Model (Vehicle-Centric)} & \textbf{I-WM (Infrastructure-Centric)} \\
\midrule
Viewpoint & Ego-vehicle perspective & Bird's-eye / multi-sensor panoramic \\
Core Strength & Fine-grained ego-surroundings modeling & Global traffic situational awareness \\
Data Foundation & $\sim$200M miles of fleet driving data & Long-term fixed-site deployment data \\
Occlusion Handling & Limited by ego viewpoint & Multi-angle occlusion completion \\
Control Dimensions & Ego driving actions & Traffic signals + multi-agent control \\
Uncertainty Modeling & No explicit reliability propagation & Quality-aware framework throughout \\
Counterfactual Scope & Single ego-vehicle ``what-if'' & Multi-agent, system-level ``what-if'' \\
Downstream Users & Single-vehicle autonomy & Traffic management, V2X, urban planning \\
Foundation Model & Genie~3 post-training & Open foundation models + roadside post-training \\
\bottomrule
\end{tabular}
\end{table}

\subsection{Temporal Depth vs.\ Spatial Breadth: The Fundamental Complementarity}
\label{sec:complementarity}

Beyond the functional differences outlined in Table~\ref{tab:comparison}, roadside and vehicle-based sensing exhibit a deeper \textbf{epistemological complementarity} rooted in their deployment geometry,one that determines what each platform can \textit{learn} about traffic dynamics.

\paragraph{Roadside LiDAR: Temporal Deep Wells.}
A fixed roadside sensor continuously observes the \textit{same} spatial region over days, weeks, or months. This produces a \textbf{temporal deep well},a dense time-series at a single location that captures the complete probability distribution of traffic behaviors, including the critical long tail of rare events (near-misses, wrong-way entries, jaywalking). Such temporal depth enables: (1)~\textit{scene normality modeling},learning what ``normal'' looks like at a specific intersection (cf.\ FRGB3D background models), so that deviations can be detected and quantified; (2)~\textit{rare-event statistics},accumulating sufficient samples of near-crash scenarios that a single passing vehicle would never encounter at the same location; and (3)~\textit{counterfactual baselines},when a near-miss occurs, the model can compare against the site's empirical behavioral distribution to ask ``what if the driver had not braked?'',a form of infrastructure-enabled counterfactual reasoning that is infeasible from a transient vehicle perspective.

\paragraph{Vehicle-Borne LiDAR: Spatial Wide Nets.}
A vehicle-mounted sensor traverses \textit{many} locations but observes each only briefly. This produces a \textbf{spatial wide net},a broad but shallow sampling of the road network that captures geometric diversity (road types, lane configurations, urban vs.\ rural), but offers only a snapshot of behavioral dynamics at any single point.

\paragraph{Implications for World Model Design.}
This complementarity has concrete architectural consequences. Roadside I-WM learns \textit{deep temporal priors},how traffic evolves over time at a specific location, including diurnal patterns, weather-induced behavioral changes, and conflict precursors. These priors are location-specific but temporally rich. Vehicle-centric WM learns \textit{broad spatial priors},what different road geometries look like, how lane configurations affect behavior, and how scenes vary across a city. These priors are spatially diverse but temporally shallow. Fused world models (Phase~III of our roadmap) can inherit temporal depth from roadside and spatial breadth from vehicles, achieving a level of environmental understanding that neither platform can reach alone.

\paragraph{Data Efficiency Argument.}
A single roadside deployment operating for one week at a busy intersection can observe thousands of unique vehicle-vehicle interactions, including dozens of conflict events. Achieving comparable behavioral diversity from a vehicle platform would require driving hundreds of kilometers through similar intersections,and even then, the vehicle would lack the repeated-observation advantage needed to estimate behavioral distributions. This \textbf{data efficiency} is a key practical motivation for infrastructure-centric world models, particularly for safety-critical applications where rare events matter most.

% ============================================================
\section{Related Work}
\label{sec:related}
% ============================================================

The proposed I-WM research vision draws on and extends several active lines of research. This section reviews the most relevant prior work across seven pillars and establishes a taxonomy of existing driving world models.

\subsection{World Models for Autonomous Driving}
\label{sec:related_wm}
The concept of \textit{world models},internal simulators that predict how the environment evolves,has gained significant traction in autonomous driving. Recent surveys provide comprehensive overviews of the field~\cite{feng2025survey, ding2025understanding, tu2025role, guan2024world}, while a dedicated review on connected automated vehicle scenarios highlights the potential of world models for cooperative perception and V2X applications~\cite{wang2025worldmodels_cav}. Representative vehicle-centric world models include GAIA-1~\cite{hu2023gaia}, which generates realistic driving videos conditioned on actions and text; Vista~\cite{gao2024vista}, a generalizable driving world model with versatile controllability; DriveDreamer~\cite{wang2024drivedreamer}, which leverages structured traffic constraints; and Epona~\cite{zhang2025epona}, an autoregressive diffusion model supporting minute-long horizon generation.

More recently, several works have pushed the frontier of unified scene understanding and generation. HERMES~\cite{zhou2025hermes} (ICCV 2025) presents a unified driving world model that simultaneously handles 3D scene understanding and future scene generation through BEV representations and LLM-based world queries, achieving state-of-the-art results on nuScenes. ResWorld~\cite{zhang2026resworld} (ICLR 2026) introduces temporal residual modeling that separates dynamic objects from static backgrounds,an idea conceptually related to our FRGB3D background subtraction but applied in latent BEV space for end-to-end planning. DriveVLA-W0~\cite{li2025drivevlaw0} (ICLR 2026) demonstrates that world models can amplify data scaling laws for autonomous driving by combining vision-language-action architectures with latent world modeling.

At the industry level, NVIDIA's Cosmos platform~\cite{nvidia2025cosmos} has established open-weight world foundation models trained on 9,000 trillion tokens from 20 million hours of real-world data, providing a general-purpose WFM that can be fine-tuned for domain-specific applications. However, \textbf{all existing driving world models,both academic and industrial,adopt an ego-vehicle perspective}, leaving infrastructure-centric world modeling as a significant and timely gap.

\subsection{4D Occupancy Forecasting}
Predicting volumetric occupancy over time is a core capability for any world model. OccWorld~\cite{zheng2024occworld} pioneered GPT-style autoregressive forecasting in 3D occupancy space, while Drive-OccWorld~\cite{yang2025driveoccworld} extended this paradigm to end-to-end planning by integrating action-conditioned generation with an occupancy-based planner. DriveWorld~\cite{min2024driveworld} proposed 4D pre-training for scene understanding via world models. These methods demonstrate that dense occupancy representations can serve as a powerful ``canvas'' for world model rollouts. Our I-WM adapts this idea to the roadside setting, where the fixed, elevated viewpoint enables more complete occupancy estimation with fewer occlusions.

\subsection{Neural Scene Reconstruction and LiDAR Generation}
Generating realistic sensor data is essential for simulation-based training. NeRF-LiDAR~\cite{zhang2024nerflidar} demonstrated that Neural Radiance Fields can produce LiDAR point clouds nearly indistinguishable from real data. In the diffusion model family, LiDAR Diffusion Models~\cite{ran2024lidar_diffusion} introduced scene-level conditional generation on KITTI, LidarDM~\cite{zyrianov2024lidardm} achieved layout-aware 4D LiDAR video generation, and RangeLDM~\cite{hu2024rangeldm} proposed fast range-view generation with structural fidelity. The Waymo World Model~\cite{waymo2026worldmodel} extends this line by jointly generating camera and LiDAR data from Genie~3 post-training. An emerging trend is \textit{multi-modal sensor generation}: as world models move toward richer sensory outputs, 4D millimeter-wave radar is gaining attention as a complementary modality. Recent surveys~\cite{fan2024radar4d_survey} document rapid progress in 4D radar perception, while learning-based radar data generation methods are beginning to close the sim-to-real gap for radar point clouds. Our roadmap builds on these advances but targets generation from the \textit{roadside} perspective, with multi-modal outputs (LiDAR + camera + radar) and the additional requirement of quality-aware (uncertainty-annotated) synthetic data.

\subsection{Physics-Informed Deep Learning and RL-Based Traffic Control}
Integrating traffic flow theory into neural networks has emerged as a promising paradigm. Raissi et al.~\cite{raissi2019pinn} established the foundational Physics-Informed Neural Networks (PINNs) framework. In transportation, Di et al.~\cite{di2023pidl_survey} surveyed PIDL architectures for traffic state estimation, proposing hybrid computational graphs that combine physics-based and data-driven components. STDEN~\cite{ji2022stden} embeds differential equation-based traffic dynamics into neural ODE solvers for flow prediction, while Zhang et al.~\cite{zhang2024pidl_traffic} developed computational graph methods that encode LWR and ARZ traffic models into deep learning pipelines. These physics-informed approaches directly inform our Phase~II design (Section~\ref{sec:phase2}) of latent dynamics with physical constraints ($\mathcal{L}_{\text{physics}}$) and counterfactual reasoning capabilities.

Complementarily, reinforcement learning (RL) for traffic signal control has matured into a rich field, with recent surveys~\cite{noaeen2025rl_tsc_review} documenting the evolution from single-agent DQN approaches to multi-agent architectures using graph neural networks and hierarchical RL for large-scale network coordination~\cite{shen2025shlight}. These RL-based signal control methods are directly relevant to our Phase~II closed-loop infrastructure simulation, where the world model serves as a ``mental simulator'' for the signal controller. Integrating a learned world model with RL-based signal optimization,replacing the traditional traffic simulator with a generative world model,represents a key innovation of the I-WM framework, enabling model-predictive control that anticipates traffic dynamics informed by Signal Phase and Timing (SPaT) data.

\subsection{Cooperative Perception and V2X Communication}
Vehicle-to-Everything (V2X) cooperative perception has become a rapidly growing field. Huang et al.~\cite{huang2025v2x_cp} provide a comprehensive survey covering agent selection, data alignment, and fusion strategies. Key datasets include DAIR-V2X~\cite{yu2022dairv2x}, V2X-Seq~\cite{yu2023v2xseq}, V2X-Real~\cite{xiang2024v2x_real}, and V2XScenes~\cite{wang2025v2xscenes} (ICCV 2025). V2X-ViT~\cite{xu2022v2xvit} introduced vision transformers for cooperative perception, and UniV2X~\cite{li2024univ2x} demonstrated end-to-end autonomous driving through V2X cooperation. Recent ICCV 2025 workshop challenges~\cite{hao2025v2xchallenge} further highlight the community's growing interest. Notably, V2X-R~\cite{huang2025v2xr} (CVPR 2025) introduced cooperative LiDAR-4D radar fusion for 3D object detection using denoising diffusion, demonstrating that 4D radar can enhance cooperative perception robustness in adverse conditions,directly supporting our multi-modal sensor strategy (Section~\ref{sec:multimodal}). While these works focus on \textit{perception-level} fusion, our Phase~III proposes a deeper form of collaboration,\textit{world model alignment},where infrastructure and vehicles share compressed latent representations of their respective world models.

\subsection{Taxonomy: Three Paradigms of Driving World Models}
\label{sec:taxonomy}

Despite the rapid proliferation of driving world models, existing literature lacks a clear taxonomy that relates their architectural choices to their downstream purposes. We identify three distinct paradigms, summarized in Table~\ref{tab:taxonomy}.

\paragraph{Paradigm A: Video Generation World Models.}
This paradigm treats world modeling as a conditional video prediction problem. Representative methods include LingBot-World~\cite{lingbot2025}, GAIA-1~\cite{hu2023gaia}, Vista~\cite{gao2024vista}, Epona~\cite{zhang2025epona}, and NVIDIA Cosmos~\cite{nvidia2025cosmos}. These models excel at visual fidelity and long-horizon generation, but operate in 2D pixel space and lack explicit 3D geometric understanding, limiting their utility for safety-critical spatial reasoning tasks.

\paragraph{Paradigm B: 3D/4D Scene Generation World Models.}
This paradigm generates or predicts directly in 3D volumetric space. DynamicCity~\cite{bian2025dynamiccity} (ICLR 2025 Spotlight) introduced HexPlane-based 4D occupancy generation with DiT-based diffusion. OccWorld~\cite{zheng2024occworld} pioneered autoregressive 3D occupancy forecasting, Copilot4D~\cite{zhang2024copilot4d} demonstrated unsupervised LiDAR world models, and HERMES~\cite{zhou2025hermes} (ICCV 2025) unifies 3D scene understanding and generation through BEV representations. HunyuanWorld~\cite{hunyuanworld2025} and UrbanWorld~\cite{shang2024urbanworld} generate explorable 3D environments. These models preserve true 3D geometry, enabling physically meaningful downstream applications.

\paragraph{Paradigm C: Decision/Planning-Oriented World Models.}
This paradigm treats the world model as an internal simulator for RL agents. CarDreamer~\cite{gao2024cardreamer} provides an open-source platform integrating DreamerV2/V3 with CARLA. Drive-WM~\cite{wang2024drivewm} (CVPR 2024) pioneered multiview forecasting with planning. DreamerV3~\cite{hafner2025dreamerv3} (Nature 2025) demonstrated mastery of diverse control tasks. ResWorld~\cite{zhang2026resworld} (ICLR 2026) bridges Paradigms B and C via temporal residuals, and DriveVLA-W0~\cite{li2025drivevlaw0} (ICLR 2026) unifies world modeling with VLA architectures.

\begin{table}[t]
\centering
\caption{Taxonomy of Driving World Model Paradigms.}
\label{tab:taxonomy}
\small
\renewcommand{\arraystretch}{1.35}
\begin{tabular}{>{\raggedright\arraybackslash}p{2.6cm} >{\raggedright\arraybackslash}p{3.2cm} >{\raggedright\arraybackslash}p{3.2cm} >{\raggedright\arraybackslash}p{4.0cm}}
\toprule
& \textbf{A: Video Generation} & \textbf{B: 3D/4D Scene Gen.} & \textbf{C: Decision/Planning} \\
\midrule
Output Space & 2D video frames & 3D occupancy / point clouds & Latent states + actions \\
Core Strength & Visual fidelity, long horizon & Geometric accuracy, physics & Sample-efficient RL \\
3D Geometry & Implicit only & Explicit & Abstract (latent) \\
Open-Source Rep. & LingBot-World, Vista, Cosmos & DynamicCity, OccWorld, HERMES, UrbanWorld & CarDreamer, DreamerV3, ResWorld, DriveVLA-W0 \\
Limitations & No true 3D reasoning & Computationally expensive & Not visually interpretable \\
\textbf{I-WM Relevance} & Reference for generation & \textbf{Primary foundation} & Phase~II control \\
\bottomrule
\end{tabular}
\end{table}

\subsection{Broader AI Landscape: JEPA, Spatial Intelligence, and VLA}
\label{sec:jepa_spatial_vla}

Beyond driving-specific world models, three broader research threads shape the conceptual foundations of our I-WM.

\paragraph{LeCun's JEPA Framework.}
Yann LeCun's Joint-Embedding Predictive Architecture (JEPA)~\cite{lecun2022jepa} proposes that world models should predict in \textit{abstract representation spaces} rather than in pixel or token spaces. This vision has materialized through I-JEPA~\cite{assran2023ijepa} for images, V-JEPA 2~\cite{bardes2025vjepa2} for video understanding and action anticipation, and LeWorldModel (LeWM)~\cite{maes2026leworldmodel}, the first JEPA that trains stably end-to-end from raw pixels using only two loss terms (a prediction loss and a Gaussian regularizer based on SIGReg~\cite{balestriero2025lejepa}), achieving 48$\times$ faster planning than foundation-model-based alternatives.

JEPA's core principle,predicting in latent space rather than reconstructing observations,directly aligns with our Phase~II latent dynamics formulation ($z_{t+1} = \mathcal{T}(z_t, a_t, \eta)$). However, JEPA methods currently operate in modality-agnostic latent spaces without explicit 3D geometry, which is insufficient for distance-based safety metrics (TTC, PET). Our I-WM bridges this gap by adopting JEPA's latent prediction philosophy within a \textit{geometrically explicit} 3D/4D representation (Paradigm~B). LeWM's SIGReg regularizer can be directly applied to stabilize training of our HexPlane latent representations, and V-JEPA 2's self-supervised video encoder offers a promising initialization strategy for the camera branch of our multi-modal roadside encoder.

\paragraph{Li Fei-Fei's Spatial Intelligence.}
Fei-Fei Li's concept of \textit{spatial intelligence}~\cite{li2025spatialintelligence} argues that AI must move beyond language to develop a grounded understanding of 3D physical environments,encompassing geometry, semantics, physics, and dynamics. World Labs' Marble model~\cite{worldlabs2025marble} operationalizes this vision by generating persistent, editable 3D environments from text, images, or video, with hybrid editing tools that decouple spatial structure from visual style.

We argue that \textbf{roadside infrastructure is the ideal platform for spatial intelligence in transportation}: the fixed bird's-eye viewpoint, multi-sensor coverage, and persistent observation provide the most complete and least occluded 3D understanding of traffic scenes. While World Labs targets general-purpose spatial AI, our I-WM represents a \textit{domain-specific spatial intelligence} optimized for transportation safety,where the temporal-depth advantage (Section~\ref{sec:complementarity}) provides behaviorally rich training data that general-purpose models cannot match.

\paragraph{Vision-Language-Action (VLA) Models and Infrastructure VLA.}
The VLA paradigm~\cite{li2025drivevlaw0} unifies visual perception, language understanding, and action generation. Current VLA models define ``action'' as ego-vehicle driving commands. We propose reconceptualizing this for the infrastructure domain as \textbf{Infrastructure VLA (I-VLA)}, where: \textit{Vision} = multi-LiDAR + camera roadside perception; \textit{Language} = natural language scene descriptions and traffic management commands; and \textit{Action} = infrastructure control decisions (signal phase optimization, dynamic speed limits, variable message signs, incident alerts). This I-VLA formulation emerges naturally from our phased roadmap and represents a novel contribution entirely absent from the current literature, where all VLA models assume an ego-vehicle agent.

\vspace{0.3cm}
\noindent\textbf{Summary of Gaps.} Despite significant progress across vehicle-centric world models, physics-informed traffic modeling, V2X cooperative perception, and broader AI paradigms including JEPA, spatial intelligence, and VLA, no existing work has combined these threads into an infrastructure-centric world model with quality-aware uncertainty propagation and V2X latent sharing. The fundamental temporal-depth advantage of fixed roadside deployment remains entirely unexploited. This paper aims to fill precisely this gap.

% ============================================================
\section{Technical Positioning and Data Strategy}
\label{sec:positioning}
% ============================================================

\subsection{I-WM Technical Positioning}

Our Infrastructure-centric World Model is positioned as \textbf{Paradigm B (3D/4D Scene Generation) as the primary foundation, augmented with elements from Paradigm C (Decision/Planning)} for closed-loop infrastructure control. This choice is driven by three requirements: (1)~roadside safety metrics (TTC, PET) require accurate 3D positions and velocities; (2)~our core data modality is multi-LiDAR point clouds, for which Paradigm~B models provide the most natural representation; and (3)~propagating spatially-varying uncertainty requires explicit 3D representations where confidence can be associated with each voxel or point.

Table~\ref{tab:opensource} identifies the open-source models most relevant to each phase of our research roadmap.

\begin{table}[t]
\centering
\caption{Open-Source Foundations for I-WM Development.}
\label{tab:opensource}
\small
\renewcommand{\arraystretch}{1.35}
\begin{tabular}{>{\raggedright\arraybackslash}p{2.8cm} >{\raggedright\arraybackslash}p{2.8cm} >{\raggedright\arraybackslash}p{3.0cm} >{\raggedright\arraybackslash}p{4.2cm}}
\toprule
\textbf{Open-Source Model} & \textbf{Venue / Year} & \textbf{Target Phase} & \textbf{Components to Adapt} \\
\midrule
DynamicCity~\cite{bian2025dynamiccity} & ICLR 2025 Spotlight & Phase I (core) & HexPlane 4D repr.; DiT diffusion; trajectory conditioning \\
OccWorld~\cite{zheng2024occworld} & ECCV 2024 & Phase I & GPT-style autoregressive occupancy forecasting \\
Copilot4D~\cite{zhang2024copilot4d} & ICLR 2024 & Phase I & Unsupervised discrete diffusion for LiDAR \\
LidarDM~\cite{zyrianov2024lidardm} & ICRA 2025 & Phase I & Layout-aware 4D LiDAR generation \\
HERMES~\cite{zhou2025hermes} & ICCV 2025 & Phase I--II & BEV tokenizer; LLM world queries \\
NVIDIA Cosmos~\cite{nvidia2025cosmos} & CES 2025 & Phase I (ref.) & Open-weight WFM; fine-tuning framework \\
CarDreamer~\cite{gao2024cardreamer} & IEEE IoT 2024 & Phase II--III & Gym interface; DreamerV2/V3; V2X sharing \\
ResWorld~\cite{zhang2026resworld} & ICLR 2026 & Phase II & Temporal residual BEV modeling \\
\bottomrule
\end{tabular}
\end{table}

\begin{figure}[t]
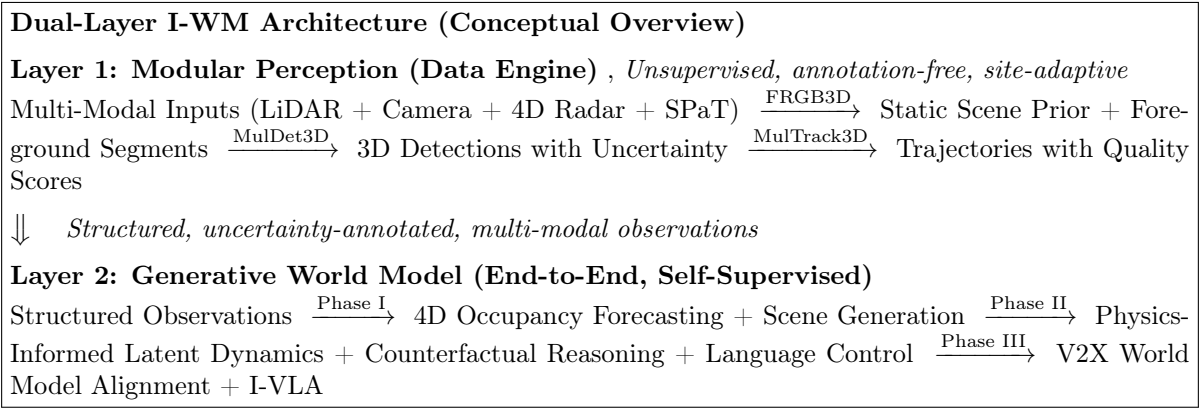

\centering
\fbox{\parbox{0.95\textwidth}{\small
\textbf{Dual-Layer I-WM Architecture (Conceptual Overview)} \\[6pt]
\textbf{Layer 1: Modular Perception (Data Engine)} , \textit{Unsupervised, annotation-free, site-adaptive} \\
Multi-Modal Inputs (LiDAR + Camera + 4D Radar + SPaT) $\;\xrightarrow{\text{FRGB3D}}\;$ Static Scene Prior + Foreground Segments $\;\xrightarrow{\text{MulDet3D}}\;$ 3D Detections with Uncertainty $\;\xrightarrow{\text{MulTrack3D}}\;$ Trajectories with Quality Scores \\[6pt]
$\big\Downarrow$ \quad \textit{Structured, uncertainty-annotated, multi-modal observations} \\[6pt]
\textbf{Layer 2: Generative World Model (End-to-End, Self-Supervised)} \\
Structured Observations $\;\xrightarrow{\text{Phase I}}\;$ 4D Occupancy Forecasting + Scene Generation $\;\xrightarrow{\text{Phase II}}\;$ Physics-Informed Latent Dynamics + Counterfactual Reasoning + Language Control $\;\xrightarrow{\text{Phase III}}\;$ V2X World Model Alignment + I-VLA
}}
\caption{Conceptual overview of the dual-layer I-WM architecture. Layer~1 provides annotation-free, multi-modal perception as a data engine; Layer~2 builds end-to-end generative world models on top of the structured representations.}
\label{fig:concept}
\end{figure}

\textbf{Core Technical Strategy.} Our I-WM adopts a dual-layer architecture (Figure~\ref{fig:concept}) that cleanly separates \textit{perception} from \textit{world modeling}. The perception layer,comprising FRGB3D (background modeling), MulDet3D (unsupervised detection), and MulTrack3D (consensus tracking),operates without manual annotation and produces structured, uncertainty-annotated representations: 3D bounding boxes, trajectories, static scene priors, and per-point reliability scores. These serve as the \textit{observation model} and \textit{data engine} for the world model layer, which employs end-to-end generative architectures trained via self-supervised objectives (predicting future occupancy from past observations,no human labels required). This separation is strategically motivated: roadside perception must be annotation-free for scalable deployment across diverse intersections, while world models benefit from end-to-end learning over the rich structured representations that the perception layer provides.

Concretely, Phase~I adapts DynamicCity's HexPlane VAE + DiT generation framework from the ego-vehicle to the roadside multi-LiDAR domain, adding quality-aware uncertainty channels, informed by HERMES's unified understanding-generation paradigm. Phase~II integrates CarDreamer's RL-based planning loop with physics-informed latent dynamics, leveraging ResWorld's temporal residual approach and LeWM's SIGReg regularizer. Phase~III develops cross-domain world model alignment protocols for V2X latent sharing, enabling roadside temporal priors to complement vehicle spatial priors.

\subsection{Data Acquisition Strategy}

Training an infrastructure-centric world model requires large-scale, multi-modal data from the roadside perspective. We adopt a \textbf{three-tier data strategy}:

\paragraph{Tier 1: Proprietary Multi-LiDAR Roadside Data.}
Our existing deployments provide multi-LiDAR synchronized point clouds with overlapping fields of view, quality-aware annotations with per-point reliability scores, 3D bounding boxes and trajectories from MulDet3D and MulTrack3D, background models from FRGB3D, and continuous multi-day recordings providing the complete behavioral distribution needed to train temporal priors.

\paragraph{Tier 2: Public Roadside and V2X Datasets.}
Key public datasets include DAIR-V2X~\cite{yu2022dairv2x} (71K frames, 28 intersections), V2X-Seq~\cite{yu2023v2xseq} (15K frames, sequential perception), TUMTraf-V2X~\cite{zimmer2024tumtrafv2x} and TUMTraf-I~\cite{zimmer2023tumtraf_i} (multi-sensor roadside benchmarks), V2X-Real~\cite{xiang2024v2x_real} (33K LiDAR frames, multi-mode cooperation), and V2XScenes~\cite{wang2025v2xscenes} (challenging multi-condition traffic).

\paragraph{Tier 3: Simulation-Generated Synthetic Data.}
CARLA-based roadside simulation provides data with perfect ground truth labels, controllable conditions, and programmable rare events. World model self-play (bootstrap) enables the trained I-WM to generate additional synthetic training data.

\subsection{Multi-Modal Sensor Strategy}
\label{sec:multimodal}

A world model's ability to understand and predict the physical world depends critically on the richness of its sensory input. We adopt a \textbf{phased multi-modal strategy} organized into three tiers of sensor modalities, reflecting both deployment readiness and value to the world model.

\paragraph{Core Modalities: Multi-LiDAR + RGB Camera.}
Multi-LiDAR point clouds form the geometric backbone of I-WM, providing precise 3D structure essential for safety metrics (TTC, PET). RGB cameras complement LiDAR with dense appearance and semantic information (lane markings, traffic signs, pedestrian intent cues). Phase~I begins as \textit{LiDAR-centric}, leveraging our existing perception pipeline, and progressively incorporates synchronized cameras for joint LiDAR-camera world model training as deployment sites are augmented.

\paragraph{Extended Modalities: 4D Radar + Signal Phase and Timing (SPaT).}
Two additional modalities offer high value with low deployment barrier:
\begin{itemize}[leftmargin=*]
    \item \textbf{4D millimeter-wave radar} provides Doppler velocity measurements that neither LiDAR nor camera directly offer,critical for the world model's dynamic prediction. Equally important, radar maintains performance in adverse weather (rain, fog, snow) where LiDAR degrades significantly, providing a natural reliability complement within our quality-aware framework. Recent work on cooperative LiDAR-4D radar fusion~\cite{huang2025v2xr} (CVPR 2025) demonstrates the feasibility of this combination in V2X settings.
    \item \textbf{Signal Phase and Timing (SPaT) data} from traffic signal controllers provides the single most important \textit{causal variable} governing intersection behavior. Without SPaT, the world model must implicitly infer signal state from vehicle behavior,an unnecessary ambiguity. With SPaT, Phase~II counterfactual reasoning can directly explore signal-conditioned scenarios (``what if the signal had turned red 5 seconds earlier?''), and Phase~II closed-loop simulation can jointly optimize signal timing and safety. SPaT data is typically available at zero marginal cost from existing signal controller infrastructure.
\end{itemize}

\paragraph{Forward-Looking Modalities: Event Cameras + Environmental Sensors.}
For longer-term exploration: (1)~\textit{event cameras} offer microsecond-level temporal resolution for capturing fast dynamics (sudden braking, collision instants) that conventional cameras miss at standard frame rates; and (2)~\textit{environmental sensors} (temperature, precipitation, visibility, road surface condition) provide conditioning variables for weather-aware world modeling,the same intersection exhibits fundamentally different behavioral distributions on dry versus icy pavement. These modalities are included in the Phase~II--III timeline as deployment infrastructure matures.

Table~\ref{tab:modality} summarizes the multi-modal strategy and its relationship to the I-WM development phases.

\begin{table}[t]
\centering
\caption{Phased Multi-Modal Sensor Strategy for I-WM.}
\label{tab:modality}
\small
\renewcommand{\arraystretch}{1.3}
\begin{tabular}{>{\raggedright\arraybackslash}p{2.2cm} >{\raggedright\arraybackslash}p{3.0cm} >{\raggedright\arraybackslash}p{3.5cm} >{\raggedright\arraybackslash}p{4.3cm}}
\toprule
\textbf{Tier} & \textbf{Modalities} & \textbf{Key Contribution to WM} & \textbf{Deployment Phase} \\
\midrule
Core & Multi-LiDAR, RGB Camera & 3D geometry, appearance, semantics & Phase I (LiDAR first, camera added progressively) \\
Extended & 4D Radar, SPaT & Doppler velocity, weather robustness, causal signal state & Phase I--II \\
Forward-Looking & Event Camera, Environmental Sensors & Microsecond dynamics, weather-conditioned behavioral priors & Phase II--III \\
\bottomrule
\end{tabular}
\end{table}

% ============================================================
\section{Research Vision}
\label{sec:roadmap}
% ============================================================

\subsection{Phase I: Generative Scene Understanding (Years 1--2)}
\label{sec:phase1}

\textit{Goal: Solving data scarcity through high-fidelity reconstruction and generation, with built-in reliability quantification. The world model layer is trained end-to-end via self-supervised objectives,predicting future scene states from past observations,without requiring manual annotation beyond what the perception layer already provides.}

\paragraph{Foundation Model Adaptation.}
Following the paradigm demonstrated by Waymo (Genie~3 $\rightarrow$ driving domain) and NVIDIA Cosmos (general WFM $\rightarrow$ domain-specific fine-tuning), we adopt a two-stage strategy: leverage open-source 3D/4D foundation models,specifically DynamicCity's HexPlane VAE and DiT diffusion framework~\cite{bian2025dynamiccity},then perform domain adaptation using our multi-LiDAR roadside datasets. The perception layer (FRGB3D, MulDet3D, MulTrack3D) automatically generates structured training data,occupancy grids, 3D bounding boxes, and trajectories,at each deployment site without human annotation, enabling the world model to be trained in a fully self-supervised manner: the ground truth for future prediction is simply the perception layer's output at future time steps. HERMES's~\cite{zhou2025hermes} approach of unifying understanding and generation through BEV representations provides a complementary paradigm for multi-task roadside scene reasoning.

\paragraph{4D Occupancy Forecasting.}
Building on OccWorld's~\cite{zheng2024occworld} autoregressive prediction paradigm, we develop algorithms to predict the volumetric occupancy of intersections in temporal sequences. The roadside viewpoint offers a key advantage: a more complete and less occluded occupancy field compared to ego-vehicle perspectives.

\paragraph{Multi-Modal Generative Synthesis.}
We employ Diffusion Models to jointly synthesize LiDAR point clouds and camera imagery from the roadside perspective, utilize Neural Radiance Fields for digital twin creation, and synthesize rare edge-case scenarios for generative data augmentation.

\paragraph{Quality-Aware Synthetic Data (Unique Contribution).}
Unlike existing world models that generate data without uncertainty information, our quality-aware framework propagates sensor-level reliability through the generation pipeline. We extend DynamicCity's HexPlane representation with additional uncertainty channels, encoding per-voxel reliability alongside semantic and geometric features.

\subsection{Phase II: Physics-Informed Predictive Dynamics (Years 2--3)}
\label{sec:phase2}

\textit{Goal: Teaching the model ``common sense,'' causality, and controllability.}

\paragraph{Latent Space Dynamics with Physical Constraints.}
\begin{equation}
    z_{t+1} = \mathcal{T}(z_t, a_t, \eta) \quad \text{subject to } \mathcal{L}_{\text{physics}}
    \label{eq:transition}
\end{equation}
We learn the transition function $\mathcal{T}$ that governs how a traffic scene evolves in a compressed latent space, with traffic flow theory and vehicle dynamics embedded as physics-informed loss functions $\mathcal{L}_{\text{physics}}$. Inspired by ResWorld's~\cite{zhang2026resworld} temporal residual approach, we develop a roadside-specific variant that leverages FRGB3D's static scene priors as an explicit background model.

\paragraph{Multi-Agent Counterfactual Reasoning.}
While Waymo's counterfactual reasoning modifies the ego-vehicle's actions, our infrastructure viewpoint enables \textit{multi-agent} counterfactual analysis. The temporal-depth advantage uniquely enables a form of counterfactual reasoning unavailable to vehicle-centric models: because the roadside sensor has observed thousands of similar scenarios, the world model can compare any specific event against the site's empirical behavioral distribution,providing a data-driven counterfactual baseline.

\paragraph{Language-Conditioned Scene Editing.}
We enable engineers and traffic planners to describe scenario modifications in natural language, generating corresponding multi-modal simulations. Beyond Waymo's control axes, we introduce infrastructure-specific axes: signal phase control, traffic volume scaling, and weather/lighting modulation.

\subsection{Phase III: Collaborative World Models for V2X (Years 3--5)}
\label{sec:phase3}

\textit{Goal: Achieving ``cognitive alignment'' between infrastructure and vehicles, merging temporal depth with spatial breadth.}

\paragraph{World Model Alignment Across Domains.}
A critical research question is how to \textit{align} infrastructure and vehicle world models in a shared latent space:
\begin{equation}
    \mathcal{L}_{\text{align}} = D_{\text{KL}}\big(q_I(z \mid x_I) \;\|\; q_V(z \mid x_V)\big) + \lambda \, \mathcal{L}_{\text{task}}
    \label{eq:align}
\end{equation}
where $q_I$ and $q_V$ are the latent encoders of the infrastructure and vehicle world models, respectively.

\paragraph{Complementary World Views: Temporal Depth Meets Spatial Breadth.}
The roadside world model provides \textit{temporal priors}: learned behavioral distributions at the current location that encode cumulative knowledge from months of observation. Connected vehicles contribute \textit{spatial context}: knowledge of road network conditions beyond the infrastructure sensor's field of view. V2I latent sharing compresses complex world states into bandwidth-efficient representations for real-time communication.

\paragraph{Multi-Agent Game Theory.}
We model competitive and cooperative behaviors of human drivers and AVs, leveraging the infrastructure's complete observability to serve as an impartial referee and information broker.

% ============================================================
\section{Leveraging Current Research Strengths}
\label{sec:strengths}
% ============================================================

This research vision builds directly upon our existing technical foundations, which constitute the \textit{perception layer} (Layer~1) of the dual-layer I-WM architecture. Table~\ref{tab:mapping} maps each component to its role as a data engine for the generative world model layer.

\begin{table}[t]
\centering
\caption{Mapping Current Research to I-WM Components.}
\label{tab:mapping}
\small
\renewcommand{\arraystretch}{1.35}
\begin{tabular}{>{\raggedright\arraybackslash}p{3.5cm} >{\raggedright\arraybackslash}p{3.8cm} >{\raggedright\arraybackslash}p{6.2cm}}
\toprule
\textbf{Existing Work} & \textbf{I-WM Role} & \textbf{Description} \\
\midrule
FRGB3D (Background Modeling) & Static Scene Prior / Temporal Normality Baseline & Provides the static world ``canvas'' and enables temporal-depth-based counterfactual reasoning. \\
MulDet3D (Multi-LiDAR Detection) & Object Initialization & Supplies detected objects as initial conditions for world model rollouts. \\
MulTrack3D / ClusterTrack3D & Dynamic Agent Modeling & Provides trajectory histories as training data for agent behavior prediction. \\
Multi-LiDAR Registration & Spatial Anchor & Ground-aligned registration ensures a consistent global frame. \\
Quality-Aware Pipeline & Uncertainty Propagation & Propagates sensor-level reliability through the entire perception-to-prediction chain. \\
\bottomrule
\end{tabular}
\end{table}

% ============================================================
\section{Broader Impact}
\label{sec:impact}
% ============================================================

A key differentiator of infrastructure-centric world models is their applicability \textit{beyond} autonomous vehicle testing. Unlike Waymo's vehicle-centric model, which primarily serves AV development, I-WM enables: (1)~\textit{proactive safety analysis},generating counterfactual near-miss scenarios to compute Surrogate Safety Measures at scale; (2)~\textit{adaptive signal control optimization},using the world model as a ``mental simulator'' for model-predictive control; (3)~\textit{urban planning and design},simulating infrastructure changes before physical construction; (4)~\textit{training data generation for agencies},providing synthetic datasets for detection system calibration; and (5)~\textit{extreme event preparedness},simulating infrastructure responses to rare events for emergency planning.

% ============================================================
\section{Conclusion}
\label{sec:conclusion}
% ============================================================

The emergence of vehicle-centric world models, alongside open platforms such as NVIDIA Cosmos, LeCun's JEPA-based latent predictive architectures, and Li Fei-Fei's spatial intelligence vision, demonstrates that world modeling is becoming a cornerstone of next-generation AI. However, the infrastructure perspective remains an untapped frontier.

Central to our approach is the recognition that roadside and vehicle-based sensing are \textbf{epistemologically complementary}: roadside systems offer \textit{temporal depth},dense, long-term behavioral observations at fixed locations,while vehicle systems offer \textit{spatial breadth},diverse sampling across large road networks. By designing the I-WM architecture to explicitly exploit this complementarity, we enable environmental understanding that neither platform can achieve alone.

Our approach,grounded in Paradigm~B foundations (DynamicCity, OccWorld, HERMES), enriched with Paradigm~C elements (CarDreamer, ResWorld), informed by JEPA's latent prediction philosophy and spatial intelligence principles, and trained on a three-tier data strategy,provides a concrete and actionable path forward. The dual-layer architecture we propose,modular, annotation-free perception as a data engine feeding end-to-end generative world models,balances the practical realities of roadside deployment (where labeled data is scarce and each site is unique) with the representational power of modern generative AI (where self-supervised learning over structured observations can capture complex traffic dynamics).

Looking further ahead, two long-term visions emerge. First, as data and compute scale, the dual-layer architecture may converge into a \textbf{fully end-to-end roadside foundation model} that jointly learns perception and world modeling from raw sensor streams,analogous to how vehicle-centric systems have evolved from modular pipelines to end-to-end architectures. Second, the natural extension of I-WM toward an \textbf{Infrastructure Vision-Language-Action (I-VLA)} framework,where roadside perception, natural language commands, and traffic control actions are unified,represents a novel research direction that could fundamentally reshape how intelligent infrastructure interacts with the transportation ecosystem. The integration of I-WM with existing vehicle-centric models through latent space alignment will unlock capabilities that neither perspective can achieve alone,moving toward a future where infrastructure does not merely observe traffic, but \textit{understands and anticipates} it.

% ============================================================
% References
% ============================================================
\bibliographystyle{plainnat}
\bibliography{main}

\end{document}